\documentclass{article}

\usepackage{microtype}
\usepackage{graphicx}
\graphicspath{{./}{appendix/}{sec/fig/}{sec/fig/Dense_Img/}}
\usepackage{subfigure}
\usepackage{booktabs} 
\usepackage{array}     
\usepackage{multirow}  
\usepackage{enumitem}  
\usepackage{placeins}  
\usepackage{stfloats}
\usepackage{hyperref}

\usepackage[preprint]{icml2026}

\makeatletter
\renewcommand{\Notice@String}{\textit{ICML 2026 Workshop on Efficient Multimodal Question Answering.}}
\makeatother

\usepackage{amsmath}
\usepackage{amssymb}
\usepackage{mathtools}
\usepackage{amsthm}

\usepackage[capitalize,noabbrev]{cleveref}

\theoremstyle{plain}

\theoremstyle{definition}

\theoremstyle{remark}

\usepackage[textsize=tiny]{todonotes}

\icmltitlerunning{LARE: Low-Attention Region Encoding for Text--Image Retrieval}

\begin{document}

\twocolumn[
\icmltitle{LARE: Low-Attention Region Encoding for Text--Image Retrieval}



\icmlsetsymbol{equal}{*}

\begin{icmlauthorlist}
\icmlauthor{Abdulmalik Alquwayfili}{sdaia}
\icmlauthor{Faisal Almeshal}{sdaia}
\icmlauthor{Jumanah Almajnouni}{sdaia}
\icmlauthor{Leena Alotaibi}{sdaia}
\icmlauthor{Faisal Alhajari}{sdaia}
\icmlauthor{Mohammed Alkhrashi}{sdaia}
\icmlauthor{Alreem Almuhrij}{sdaia}
\icmlauthor{Abdullah Aldwyish}{sdaia}
\icmlauthor{Raied Aljadaany}{sdaia}
\icmlauthor{Huda Alamri}{sdaia}
\icmlauthor{Muhammad Kamran J. Khan}{sdaia}
\end{icmlauthorlist}

\icmlaffiliation{sdaia}{Saudi Data and Artificial Intelligence Authority (SDAIA), Riyadh, Saudi Arabia}

\icmlcorrespondingauthor{Abdulmalik Alquwayfili}{aalquwayfili@ncai.gov.sa}

\icmlkeywords{Machine Learning, ICML}

\vskip 0.3in
]



\printAffiliationsAndNotice{}  

\begin{abstract}
Image retrieval in crowded scenes is particularly challenging due to the salience bias of conventional visual encoders, which tend to focus on dominant objects while neglecting low-attention regions that are often crucial for fine-grained retrieval. We propose \textbf{LARE}\footnote{Code: \href{https://github.com/AbdulmalikDS/LARE}{github.com/AbdulmalikDS/LARE}} (Low-Attention Region Encoding), a framework that explicitly models these overlooked regions. LARE adopts a dual-encoding strategy that encodes low-attention regions of an image and the full image in parallel, leading to more diverse and informative image embeddings. 
To evaluate image retrieval performance in challenging crowded scenes, we introduce \textbf{Dense-Set}\footnote{Data: \href{https://huggingface.co/datasets/AbdulmalekDS/Dense-Set}{huggingface.co/AbdulmalekDS/Dense-Set}}, a challenging subset derived from COCO and Flickr30K. In this subset, images are re-captioned to provide richer descriptions of low-attention or previously overlooked regions. This dataset highlights the limitations of existing retrieval models and enables a more rigorous evaluation under densely crowded scene conditions.
Experimental results demonstrate that the proposed framework improves retrieval performance by preserving subtle, non-dominant visual cues within the shared latent space.
\end{abstract}

\section{Introduction}

\begin{figure}
    \centering
\includegraphics[width=0.9\columnwidth]{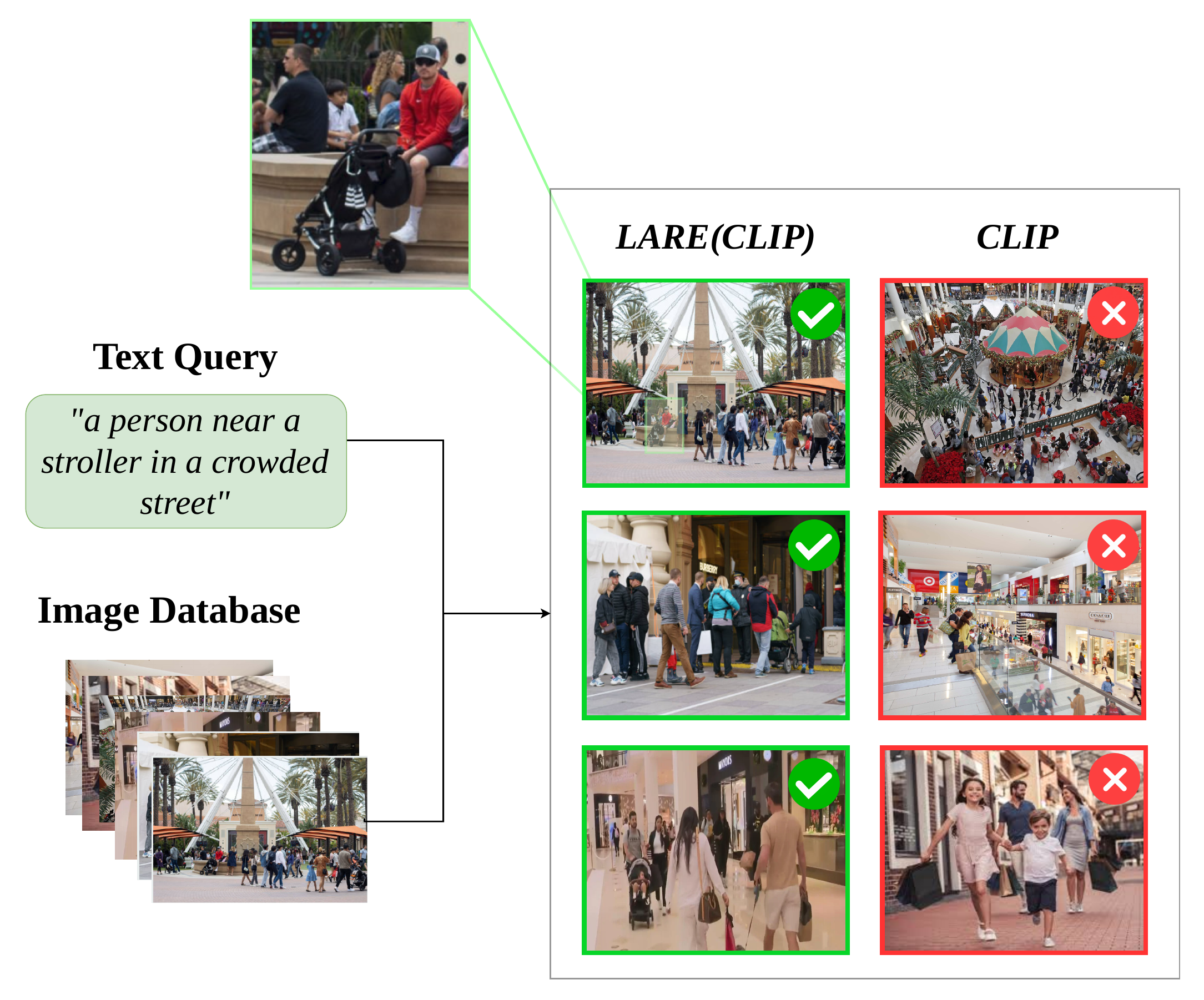}
\caption{\textbf{Fine-grained retrieval in dense scenes.} For the query \textit{``a person near a stroller in a crowded street''}, LARE retrieves results that preserve the stroller-related local cue, while CLIP tends to favor globally similar crowded scenes. Green checks indicate relevant matches; red crosses indicate mismatches.}
    \label{fig:teaser}
\end{figure}

Text-to-image retrieval retrieves images from large collections that best match a natural-language query. This capability is central to many real-world applications, including multimedia search engines, content recommendation systems, digital asset management, and large-scale visual indexing for web platforms. More broadly, cross-modal retrieval enables intuitive natural-language interaction with visual data and has become a key component in modern multimodal AI systems.~\cite{radford2021learning,jia2021scaling,yao2021filip,gao2022pyramidclip,li2021align,luo2022clip4clip,bain2021frozen,ma2022xclip,gorti2022xpool}.

Recent advances in large-scale vision–language pretraining have significantly improved cross-modal retrieval by learning shared embedding spaces in which images and text can be compared directly. Contrastive models such as CLIP~\cite{radford2021learning} and ALIGN~\cite{jia2021scaling} learn aligned visual and textual representations using massive image–text datasets, enabling strong zero-shot transfer across many tasks without task-specific training. In these models, an image encoder and a text encoder project inputs from each modality into a common embedding space, and retrieval is performed by ranking the similarity between their representations. This paradigm has become the dominant approach for cross-modal retrieval and underlies many modern multimodal systems~\cite{radford2021learning,jia2021scaling,li2022blip,zhai2023siglip,huang2021uniter,chen2020uniter,pmlr-v139-kim21k}.

Despite their success, current vision-language encoders mainly rely on a \emph{global image embedding} that summarizes the entire image into a single representation. Although effective for many queries, this representation often emphasizes the most visually salient objects or scene context while underrepresenting smaller or less prominent elements. As a result, retrieval models may overlook visually relevant cues that occupy only a small portion of the image. This limitation is particularly evident in dense scenes with many objects, where correct retrieval may depend on attributes or objects that are not dominant in the global representation. Previous work has shown that vision-language models can struggle to localize fine-grained visual evidence and often prioritize coarse scene semantics over detailed object-level information~\cite{wang2023sclip}.

In this work, we address this limitation by recovering information from image regions that receive little attention in the global representation. Our key observation is that transformer-based vision encoders implicitly encode spatial attention signals that reveal which regions contribute less to the final embedding. Rather than relying solely on the global representation, we exploit these signals to identify under-attended regions that may contain discriminative visual cues relevant to the query. 

We propose Low-Attention Region Encoding (LARE), a training-free framework that augments standard dual-encoder retrieval models with region-level evidence. Given an input image, LARE extracts low-attention regions from the encoder’s attention maps and re-encodes them to complement the global image embedding. During retrieval, the similarity between the text query and both global and regional representations is evaluated using a confidence-gated scoring mechanism.

To evaluate retrieval under challenging conditions, we introduce Dense-Set, a curated subset of COCO~\cite{lin2014microsoft} and Flickr30K~\cite{young2014image} that emphasizes crowded scenes and rare objects. The dataset contains images with many detected objects and at least one rare object instance, along with re-captioned descriptions that highlight these underrepresented elements.

Experiments show that LARE consistently improves retrieval performance in dense scenes while preserving the ranking behavior of the original encoder on standard benchmarks, without requiring additional training, parameters, or architectural modifications.

Our contributions can be summarized as follows:

\begin{itemize}
\item We propose \textbf{LARE}, a training-free retrieval framework that augments global image embeddings with region-level representations extracted from low-attention areas.

\item We introduce \textbf{Dense-Set}, a curated benchmark designed to evaluate retrieval performance in crowded scenes containing rare or visually subordinate objects.

\item We conduct extensive experiments and ablation studies demonstrating consistent improvements on dense retrieval benchmarks across multiple backbone encoders while preserving performance on standard datasets.
\end{itemize}

The remainder of the paper is organized as follows. Section~\ref{sec:related-work} reviews related work. Section~\ref{sec:denseset} introduces the Dense-Set and its construction pipeline. Section~\ref{sec:method} presents the proposed LARE retrieval framework. Section~\ref{sec:results} reports experimental results and analysis on both standard benchmarks and Dense-Set. Finally, Section~\ref{sec:conclusion} concludes the paper.

\section{Related Work}
\label{sec:related-work}

This work is related to research on text-to-image retrieval using vision–language models, methods for fine-grained image–text alignment, and approaches to retrieval in dense, visually complex scenes.

\begin{figure*}
    \centering
    \includegraphics[width=\linewidth]{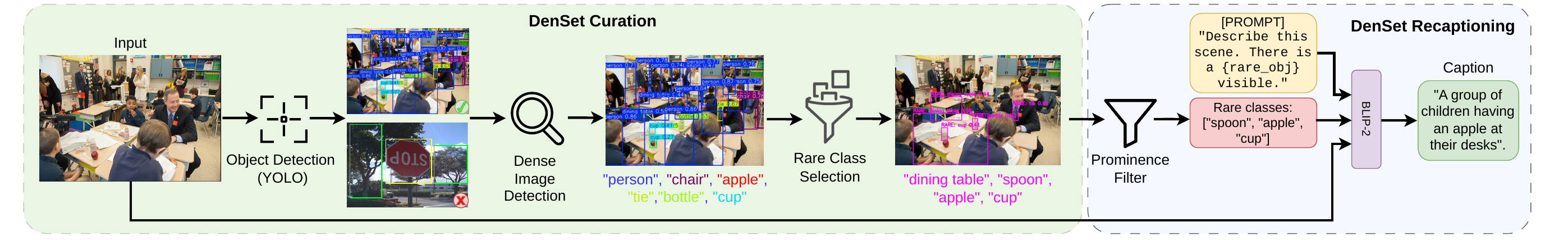}
    \caption{Dense-Set curation pipeline. We first detect objects with YOLO and rank images by total object count, retaining the top 10\% as the \emph{High-Density Subset} (dense candidate pool). We then apply rare-class filtering and keep images containing at least one single-instance class to form the final Dense-Set.}
    \label{fig:denset_curation}
\end{figure*}

\subsection{Text-to-Image Retrieval}

Text-to-image retrieval aims to retrieve images that match a natural language query, and it is a fundamental task in vision--language understanding~\cite{radford2021learning,jia2021scaling,li2022blip,zhai2023siglip,huang2021uniter,chen2020uniter,pmlr-v139-kim21k}. Early approaches learned joint embedding spaces using convolutional neural networks for visual encoding and recurrent networks for text representation~\cite{donahue2014decaf, sharif2014cnn}. More recently, large-scale vision--language pretraining has significantly improved retrieval performance by leveraging massive collections of image--text pairs~\cite{radford2021learning,li2022blip,zhan2025elip}.

Dual-encoder architectures have become the dominant paradigm for this task. Models such as CLIP and ALIGN learn aligned image and text representations using contrastive learning over large-scale datasets, enabling strong zero-shot retrieval performance across multiple benchmarks~\cite{radford2021learning,jia2021scaling}. In these models, the image and text encoders independently project each modality into a shared embedding space, allowing efficient similarity computation and scalable retrieval. Subsequent works have further improved representation quality and training efficiency. For example, BLIP introduces bootstrapped caption generation to enhance multimodal representation learning~\cite{li2022blip}, while SigLIP replaces the traditional softmax contrastive loss with a sigmoid loss to improve scalability and training stability~\cite{zhai2023siglip}. 

Despite their strong performance, dual-encoder retrieval models typically rely on a \emph{global image embedding} that summarizes the entire image into a single vector. While effective for many queries, such representations may underrepresent localized visual evidence when relevant objects occupy small or visually subordinate regions within the image.

\subsection{Fine-Grained Vision--Language Alignment}

To address the limitations of global representations, several works explore fine-grained alignment between image regions and textual tokens. FILIP introduces a late-interaction mechanism that computes token-level similarity between image patches and textual tokens, enabling finer-grained cross-modal alignment while maintaining efficient inference~\cite{yao2021filip}. PyramidCLIP further improves alignment by introducing hierarchical feature representations that capture visual semantics at multiple levels of granularity~\cite{gao2022pyramidclip}. 

Another line of work focuses on region-level representations. RegionCLIP extends contrastive language-image pretraining to region-based representations, enabling alignment between t\textbf{e}xtual concepts and localized image regions~\cite{zhong2021regionclip}. More recently, methods such as ELIP introduce lightweight text-guided visual prompts that condition the image encoder on the query, improving retrieval performance without retraining large backbone models~\cite{zhan2025elip}. 

While these approaches improve fine-grained alignment, many require additional training, architectural modifications, or query-conditioned representations, thereby increasing computational complexity. Unlike prior approaches that require retraining or query-conditioned encoders, our method augments global representations with region-level embeddings extracted at inference time, thereby improving retrieval in dense scenes while preserving the efficiency of dual-encoder architectures.

\subsection{Retrieval in Dense and Complex Scenes}

Text-to-image retrieval becomes particularly challenging in crowded scenes and long-tail object distributions, where relevant evidence may correspond to small or rare objects. Datasets such as COCO and Flickr30K contain complex scenes with multiple objects, occlusions, and visual clutter, making global image representations insufficient for capturing fine-grained attributes~\cite{lin2014microsoft, plummer2015flickr30k}. In such scenarios, correct retrieval may depend on localized visual cues that are not dominant within the scene. To address this, prior work has explored combining global and local representations, for example by leveraging local features to refine global similarity rankings~\cite{aiger2025global}.

Recent studies have also shown that attention maps produced by vision transformers encode implicit spatial signals that indicate which regions contribute most to the final representation. These signals have been used for interpretability and weak localization tasks, revealing how visual transformers allocate attention across spatial regions. Concurrent work explores a related inverse-attention idea for video retrieval~\cite{lookbeyond2026}, fusing regional and global scores via a hard maximum; in contrast, LARE targets image retrieval and introduces confidence-gated fusion together with the curated Dense-Set benchmark.

Motivated by these observations, our work leverages the internal attention structure of vision transformers to identify \emph{low-attention regions} that may contain underrepresented visual evidence.

\section{Dense-Set Dataset}
\label{sec:denseset}

To evaluate the proposed methodology, we construct \textbf{Dense-Set}, a curated benchmark of visually dense scenes. 
The goal is to create a challenging evaluation subset containing crowded images with multiple object instances and underrepresented classes. 
To this end, we develop an automated pipeline, illustrated in Figure~\ref{fig:denset_curation}. 
In the following subsections, we describe the main stages of this pipeline.

\begin{table*}[!t]
\centering
\caption{Examples from Dense-Set with rewritten captions highlighting rare or low-attention objects for more challenging dense-scene evaluation.}
\footnotesize
\setlength{\tabcolsep}{3pt}
\renewcommand{\arraystretch}{0.9}
\resizebox{\textwidth}{!}{
\begin{tabular}{
m{0.08\textwidth} 
>{\centering\arraybackslash}m{0.23\textwidth}
>{\centering\arraybackslash}m{0.23\textwidth}
>{\centering\arraybackslash}m{0.23\textwidth}
>{\centering\arraybackslash}m{0.23\textwidth}}
\toprule
\textbf{Dataset} 
& \textbf{COCO} 
& \textbf{COCO} 
& \textbf{Flickr30K}  
& \textbf{Flickr30K}  \\
\midrule

Image &
\includegraphics[width=\linewidth]{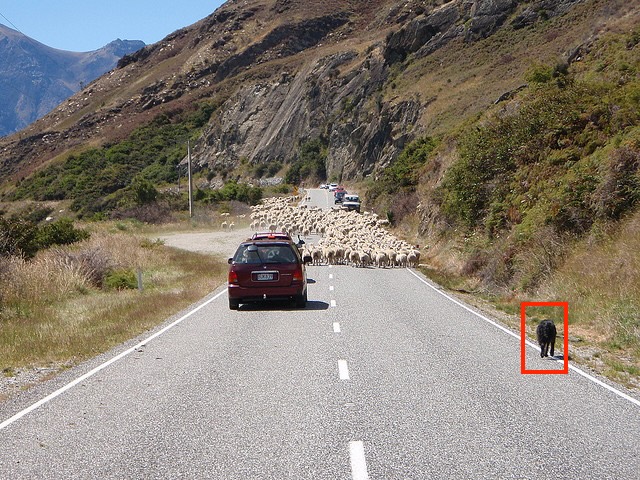} &
\includegraphics[width=\linewidth]{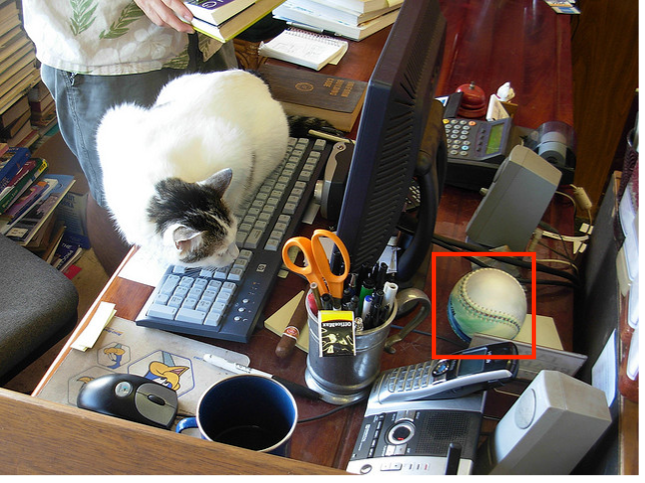} &
\includegraphics[width=\linewidth]{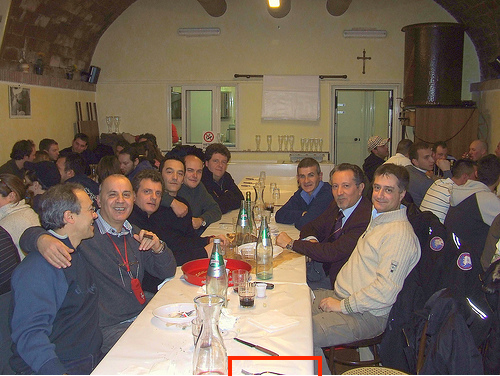} &
\includegraphics[width=\linewidth]{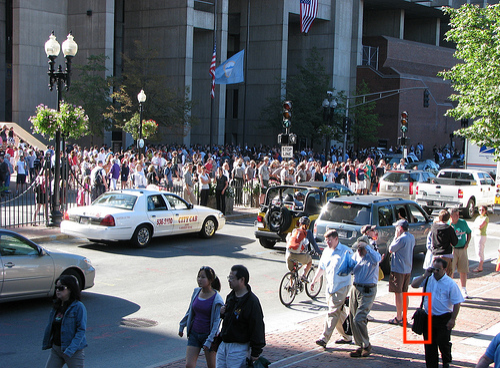} \\
\midrule

\textbf{Original Caption} &
Car driving down a road behind a lot of sheep. &
A cat lying down on a desk by a computer keyboard. &
A group of men wearing sweaters are dining in a hall. &
A crowd of people is standing outside next to a street. \\
\midrule

\textbf{Rare Class} &
Dog &
Sports ball &
Fork &
Handbag \\
\midrule

\textbf{Rewritten Caption} &
A photo of a dog standing on the side of a road with a herd of sheep. &
A sports ball sitting on top of a desk. &
A fork placed in the middle of a group of men sitting at a table. &
A handbag on the ground in front of a crowd of people. \\
\bottomrule
\label{DenseSet_examples}
\end{tabular}%
}
\end{table*}

\subsection{Dense-Set Construction}
\label{sec:denseset_construction}

This stage of the pipeline, illustrated in the first half of Figure~\ref{fig:denset_curation}, focuses on identifying densely populated images that contain underrepresented object instances. We begin by processing all images from the COCO~\cite{lin2014microsoft} and Flickr30K~\cite{young2014image} test splits using a YOLO object detector~\cite{bochkovskiy2020yolov4}. 
For each image, the detector outputs bounding boxes and class predictions, from which we compute three image-level statistics: $(i)$ the total number of detected objects, $(ii)$ the number of unique object categories, and 
$(iii)$ per-class instance frequencies.

To construct the dense candidate pool, images are ranked in descending order by total object count, and the top \textbf{10\%} are selected. 
This step favors crowded scenes with high object density and diverse visual content. Within this dense candidate set, we identify \emph{rare classes} at the image level, defined as object categories that appear exactly once in a given image. 
In crowded scenes, such single-instance categories often correspond to small or low-salience objects that are easily overlooked by global representations.

The final Dense-Set subset consists of images that $(1)$ belong to the dense candidate pool and $(2)$ contain at least one rare-class instance. 
This selection strategy yields a benchmark with significantly higher object density and class diversity than the original splits, thereby creating a more challenging setting for fine-grained text-to-image retrieval.

\begin{table}[!t]
\centering
\caption{Stage-wise statistics of Dense-Set curation for COCO and Flickr30K}

\footnotesize
\setlength{\tabcolsep}{6pt}
\resizebox{\columnwidth}{!}{%
\begin{tabular}{ll rcc}
\toprule
\textbf{Dataset} & \textbf{Split} & \textbf{\# Images} & \textbf{Avg. Objects} & \textbf{Avg. \# Classes} \\
\midrule
\multirow{3}{*}{COCO} 
& Original Test Set   & 40,504 & 6.71& 2.85\\
& High-Density Subset& 4,050& 21.63& 4.82\\
& Dense-Set           & 3,089& 21.63& 5.47\\
\midrule
\multirow{3}{*}{Flickr30K}& Original Test Set   & 31,783 & 6.73& 2.48\\
& High-Density Subset& 3,178& 19.40& 4.38\\
& Dense-Set           & 2,477& 19.55& 4.85\\
\bottomrule
\end{tabular}%
}
\label{tab:dataset_stats}
\end{table}

Table~\ref{tab:dataset_stats} summarizes the three stages shown in Figure~\ref{fig:denset_curation}: the Original Test Set, the High-Density Subset (top 10\% by object count), and the final Dense-Set after rare-class filtering. For each stage, we report the number of images, the average number of detected objects per image, and the average number of object classes.
The final curated Dense-Set contains images with substantially more objects and a broader set of object categories compared to the original splits. These characteristics make Dense-Set particularly suitable for evaluating retrieval models in visually dense environments, where important objects may appear in low-attention regions and are more likely to be overlooked by standard global representations.

\begin{figure*}[t]
  \centering
\includegraphics[width=\textwidth,height=0.5\textheight,keepaspectratio]{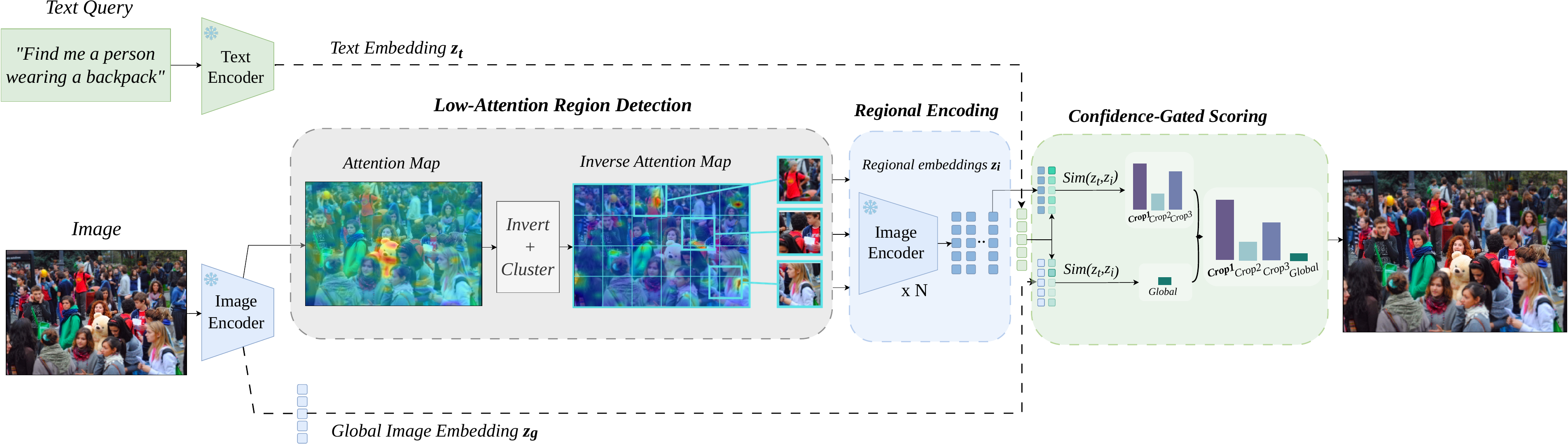}
  \caption{%
    LARE pipeline: A single forward pass produces both a global image embedding and a spatial attention map.
    Inverting the attention map highlights under-attended regions, which are clustered into candidate crops and then re-encoded independently.
    A confidence gate determines whether regional evidence should be used to adjust the final retrieval score.}
  \label{fig:Pipeline}
\end{figure*}

\subsection{Dense-Set Re-captioning}

The second stage of the pipeline, illustrated in the second half of Figure~\ref{fig:denset_curation}, focuses on regenerating captions for the curated Dense-Set images. 
The goal of this re-captioning step is to produce more challenging textual descriptions that explicitly emphasize low-attention regions, i.e., rare-class instances. 
In contrast, the original dataset captions typically describe the dominant scene context and often overlook small or underrepresented objects.
For each image in Dense-Set, we first filter rare-class detections whose bounding boxes occupy a large fraction of the image area (e.g., greater than 15\%). Such instances are likely to correspond to visually dominant objects rather than genuinely low-salience elements. 
This filtering ensures that the captioning process focuses on secondary or background objects that are more likely to be ignored by global visual representations.
The rare-class-filtered labels are then used as guidance for a vision-language model (BLIP-2). 
Specifically, we prompt the model to use class-aware templates \textit{(e.g., ``a photo of a [class]'')} to encourage explicit mention of these underrepresented objects in the generated description.
The model takes both the image and the guided prompt as input and outputs a single caption in the standard COCO format.
By shifting the caption focus from general scene-level descriptions to fine-grained object-level details, this re-captioning process produces a more demanding evaluation setting for text-to-image retrieval in dense scenes.

Examples of the curated Dense-Set and their rewritten captions are shown in Table~\ref{DenseSet_examples}. 
For each image from COCO and Flickr30K, we identify a rare or low-attention class and rewrite the original caption to explicitly describe the overlooked object. 
This shifts the textual focus from general scene context to fine-grained object-level details, thereby making dense-scene retrieval evaluation more challenging.

\section{Methodology}
\label{sec:method}

We introduce Low-Attention Region Encoding (LARE), a training-free framework that enhances visual semantic search by recovering information from regions typically underemphasized by standard vision encoders. Our approach follows a three-stage pipeline illustrated in Figure~\ref{fig:Pipeline}: (1) Low-Attention Region Detection, (2) Regional Encoding, and (3) Confidence-Gated Scoring.

\subsection{Low-Attention Region Detection}

The first stage identifies non-dominant visual cues by analyzing the internal self-attention signals of a frozen vision encoder. Given an input image $I$, we extract the self-attention tensor from an intermediate layer $\ell$. For each head $h$, let $\mathbf{A}^{(h)} \in \mathbb{R}^{HW \times HW}$ denote the patch-to-patch attention matrix.

We quantify the amount of attention each patch $i$ receives from all other patches by calculating the column-wise sum:

\begin{equation}
    a_i^{(h)} = \sum_j A^{(h)}_{j,i}, \qquad i \in \{1,\dots,HW\}
\end{equation}

Each map $a^{(h)}$ is reshaped to a spatial grid, min-max normalized, and averaged across the top-$k$ heads (selected by spatial variance) to form a mean attention map $\bar{\mathbf{A}}$. We then derive an inverse-attention map:

\begin{equation}
    \mathbf{M} = \mathbf{1} - \bar{\mathbf{A}}
\end{equation}

where high values in $\mathbf{M}$ highlight patches that consistently receive minimal attention. We apply a sliding window and non-maximum suppression (NMS) on $\mathbf{M}$ to generate a set of $N$ candidate regions, $\mathcal{R}=\{r_1,\dots,r_N\}$. We analyze sensitivity to $N$ in Appendix~\ref{sec:appendix_hyperparams}, Figure~\ref{fig:hyperparam_sensitivity}.

\subsection{Regional Encoding}

The second stage encodes the image regions generated in the previous stage.

\begin{equation}
    \mathbf{z}_i = f_v(r_i), \quad i = 1, \dots, N
\end{equation}

This produces a set of regional feature vectors $\{\mathbf{z}_1, \dots, \mathbf{z}_N\}$. Because the encoder weights are shared, these regional embeddings reside in the same feature space as the global representation, allowing for direct comparison with text embeddings without additional training.

\begin{table*}[!t]
\centering
\caption{Zero-shot retrieval performance of baseline models and LARE pipeline on COCO and Flickr30K, along with their Dense-Set variants.}
\label{tab:results}
\footnotesize
\setlength{\tabcolsep}{4pt}
\begin{tabular}{l l ccc ccc ccc ccc}
\toprule
\textbf{Model} & \textbf{ViT} 
& \multicolumn{3}{c}{\textbf{COCO}} 
& \multicolumn{3}{c}{\textbf{Flickr30K}} 
& \multicolumn{3}{c}{\textbf{COCO-Dense}} 
& \multicolumn{3}{c}{\textbf{Flickr30K-Dense}} \\ 
\cmidrule(lr){3-5} \cmidrule(lr){6-8} \cmidrule(lr){9-11} \cmidrule(lr){12-14}
 & & R@1 & R@5 & R@10 
   & R@1 & R@5 & R@10
   & R@1 & R@5 & R@10
   & R@1 & R@5 & R@10 \\ 
\midrule
CLIP \cite{radford2021learning} & L/14 
& 36.10 & 61.10 & 71.44 & 65.00 & 88.00 & 92.62 & 17.79 & 35.85 & 45.11 & 3.48 & 11.97 & 16.33 \\
SigLIP \cite{zhai2023siglip} & So/14 
& 54.24 & 76.78 & 84.21 & 82.94 & 96.08 & 98.00 & 26.61 & 46.31 & 55.22 & 5.05 & 15.50 & 20.96 \\
SigLIP 2 \cite{tschannen2025siglip} & So/16 
& 56.55 & 78.75 & 85.95 & 83.72 & 96.34 & 98.32 & 27.56 & 47.56 & 56.73 & 5.12 & 16.47 & 21.80 \\
\midrule
\textbf{LARE (CLIP)} & L/14 
& 36.10 & 61.10 & 71.44 & 65.00 & 88.00 & 92.62 & 22.97 & 42.10 & 52.03 & 9.73 & 16.63 & 20.40 \\
\textbf{LARE (SigLIP)} & So/14 
& 54.26 & 76.80 & 84.24 & 82.94 & 96.12 & 98.00 & 29.94 & 50.17 & 59.26 & 12.33 & 19.87 & 24.10 \\
\textbf{LARE (SigLIP 2)} & So/16 
& \textbf{56.56} & \textbf{78.78} & \textbf{85.97} & \textbf{83.76} & \textbf{96.38} & \textbf{98.34} & \textbf{31.00} & \textbf{51.45} & \textbf{60.67} & \textbf{13.28} & \textbf{21.11} & \textbf{25.10} \\
\bottomrule
\end{tabular}
\end{table*}

\subsection{Confidence-Gated Scoring}
Finally, we integrate the global and regional information to compute a comprehensive retrieval score. While prior work fuses regional and global signals via a hard maximum~\cite{lookbeyond2026}, this can amplify spurious regional matches when the global embedding is already well-aligned. We instead introduce a confidence-gated fusion that defers to the global score when the model is confident, and only blends in regional evidence otherwise. First, we obtain the global image embedding $\mathbf{z}_g = f_v(I)$ and the text query embedding $\mathbf{z}_t = f_t(T)$. We define the global similarity as $s_g = \text{sim}(\mathbf{z}_t, \mathbf{z}_g)$ and the strongest regional match as $s_r = \max_i \text{sim}(\mathbf{z}_t, \mathbf{z}_i)$.
To ensure robustness against regional noise, we gate the contribution of the regions based on the model's confidence in the global match. If $s_g$ exceeds a confidence threshold $\tau$, the final score remains $S = s_g$. If $s_g < \tau$ and a region outperforms the global match ($s_r > s_g$), we interpolate toward the regional score:
\begin{equation}
\label{eq:gate}
    \alpha = \min\bigl(2(s_r - s_g), 0.5\bigr), \qquad S = (1-\alpha)s_g + \alpha s_r
\end{equation}
where $\tau = 0.25$. We analyze the sensitivity to $\tau$ in Appendix~\ref{sec:appendix_hyperparams}, Figure~\ref{fig:hyperparam_sensitivity}. This fusion logic ensures that regional evidence effectively ``rescues'' the ranking when the global embedding is insufficient, particularly in dense scenes targeting non-salient objects.

\section{Results and Analysis}
\label{sec:results}

We evaluate LARE in a zero-shot image retrieval setting, where no additional training or fine-tuning is performed on the target benchmarks. Given a textual query, the task is to retrieve the most semantically aligned image from a candidate set. We compare the performance of LARE against several state-of-the-art vision–language retrieval models, including CLIP~\cite{radford2021learning}, SigLIP~\cite{zhai2023siglip}, and SigLIP~2~\cite{tschannen2025siglip}. Evaluation is conducted on COCO~\cite{lin2014microsoft} and Flickr30K~\cite{young2014image}, as well as their Dense-Set variants designed to emphasize crowded scenes and rare objects. Performance is reported using Recall@K metrics (R@1, R@5, R@10).

\subsection{Zero-Shot Retrieval Results}

\paragraph{Performance on standard datasets:} As shown in Table~\ref{tab:results}, the first two column groups (COCO and Flickr30K) report results on standard benchmark splits. On these datasets, LARE maintains performance comparable to the underlying backbone models, with differences being marginal across all Recall@K metrics. This near-zero change is \emph{by design} rather than a lack of benefit: the confidence gate (Eq.~\ref{eq:gate}) defers entirely to the global score whenever the global match is already confident, which holds for the large majority of standard-split queries whose captions describe dominant scene content. The intended behavior is therefore a no-regression guarantee on the common case, with region evidence activated only where the global embedding is insufficient. The fine-grained regime where this occurs is exactly what Dense-Set isolates, and the consistent gains there across three backbones indicate the benefit is a property of the encoder's salience bias rather than an artifact of any single split.
\paragraph{Performance on Dense-Set:} In contrast, the last two columns of Table~\ref{tab:results} (COCO-Dense and Flickr30K-Dense) demonstrate substantial gains on the curated Dense-Set benchmarks. On COCO-Dense, LARE improves R@1 by +5.18 points (29\% relative improvement) for CLIP, +3.33 points (12.5\%) for SigLIP, and +3.44 points (12.5\%) for SigLIP~2. On Flickr30K-Dense, the gains are even more pronounced: +6.25 points (180\% relative improvement) for CLIP, +7.28 points (144\% relative improvement) for SigLIP, and +8.16 points (159\% relative improvement) for SigLIP~2.

These results show that while LARE preserves performance on standard benchmarks, it delivers large and consistent improvements in dense-scene retrieval scenarios, particularly where relevant objects are rare, small, or visually subordinate.

\paragraph{Cross-Backbone Generalization:} The consistent improvement across diverse architectures (from CLIP to SigLIP~2) demonstrates that LARE operates as a general, plug-and-play inference refinement. It complements even the strongest modern encoders, suggesting that "salience bias" is a fundamental characteristic of global embeddings that persists despite scaling.

\begin{figure*}[!t]
\centering

\newlength{\imgW}
\newlength{\imgH}
\setlength{\imgW}{0.22\textwidth}
\setlength{\imgH}{0.16\textwidth}

\newcommand{\rnk}[1]{\raisebox{0.5\imgH}{\large\bfseries #1}}

\setlength{\tabcolsep}{2pt}
\renewcommand{\arraystretch}{1.05}

\begin{tabular}{@{} c c c @{\hspace{6pt}\vrule width 0.6pt\hspace{6pt}} c c @{}}

~ & 
\multicolumn{2}{c}{\parbox{0.44\textwidth}{\centering\small\itshape
Query: ``A cyclist wearing a backpack next to a train station''}} &
\multicolumn{2}{c}{\parbox{0.44\textwidth}{\centering\small\itshape
Query: ``A person carrying a red bag in a busy outdoor market''}} \\[6pt]

{\scriptsize\bfseries Rank} &
{\scriptsize\bfseries Baseline} &
{\scriptsize\bfseries LARE} &
{\scriptsize\bfseries Baseline} &
{\scriptsize\bfseries LARE} \\[4pt]

\rnk{1} & \includegraphics[width=\imgW,height=\imgH]{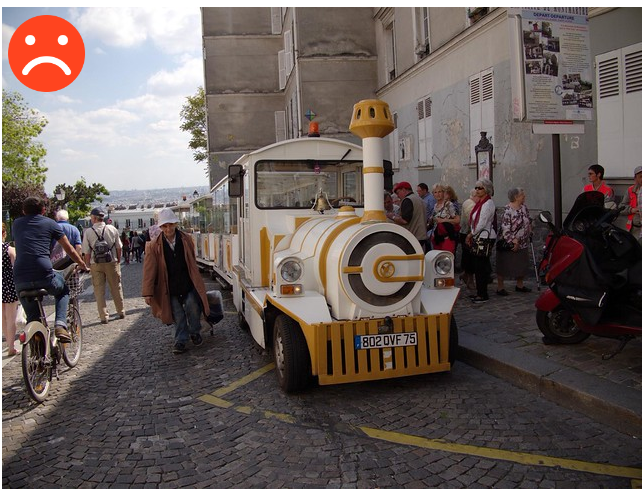} & \includegraphics[width=\imgW,height=\imgH]{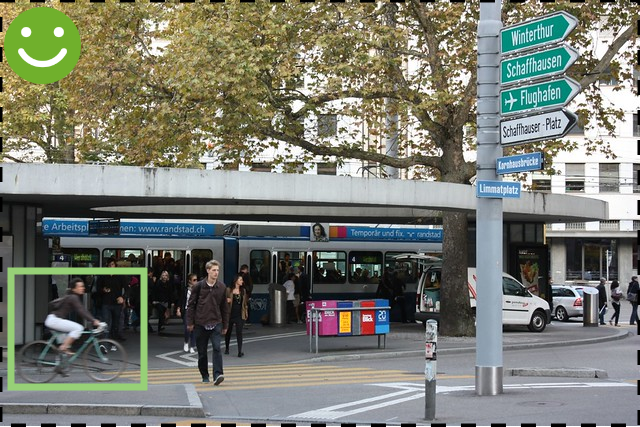} & \includegraphics[width=\imgW,height=\imgH]{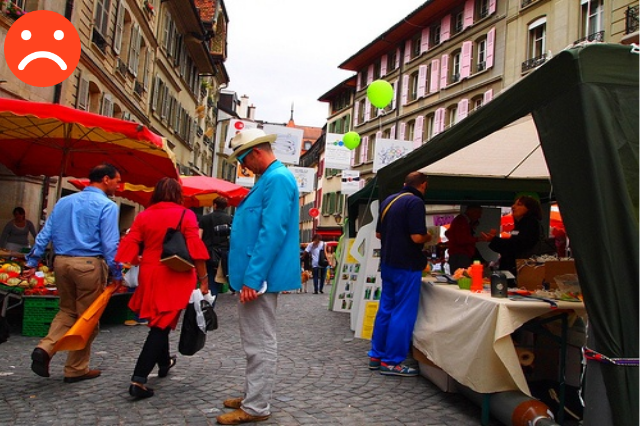} & \includegraphics[width=\imgW,height=\imgH]{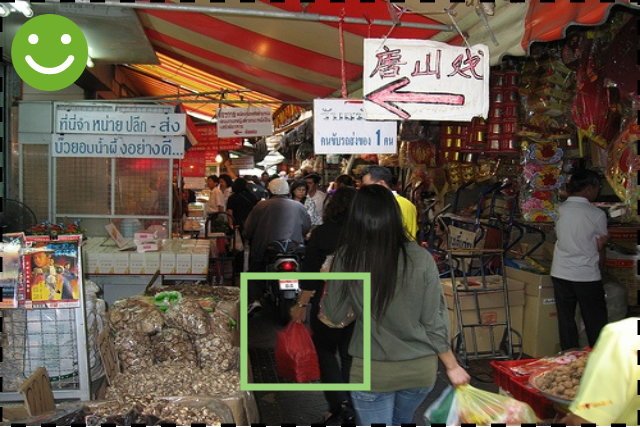} \\[6pt]
\rnk{2} & \includegraphics[width=\imgW,height=\imgH]{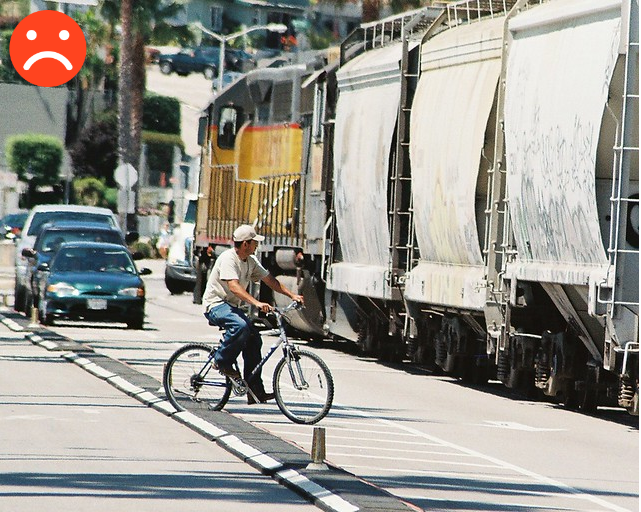} & \includegraphics[width=\imgW,height=\imgH]{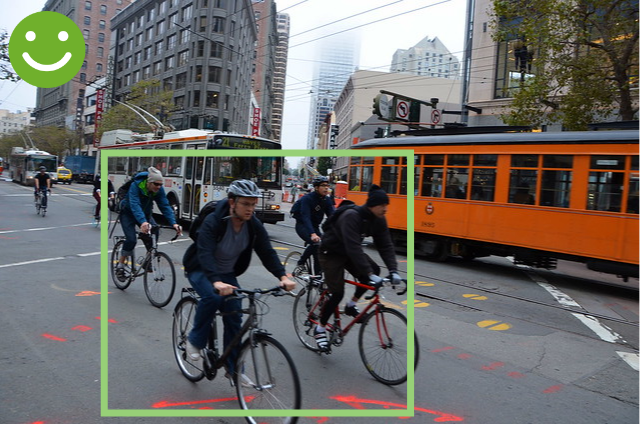} & \includegraphics[width=\imgW,height=\imgH]{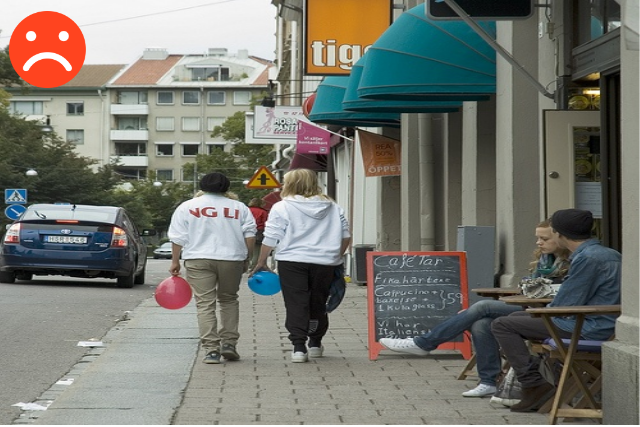} & \includegraphics[width=\imgW,height=\imgH]{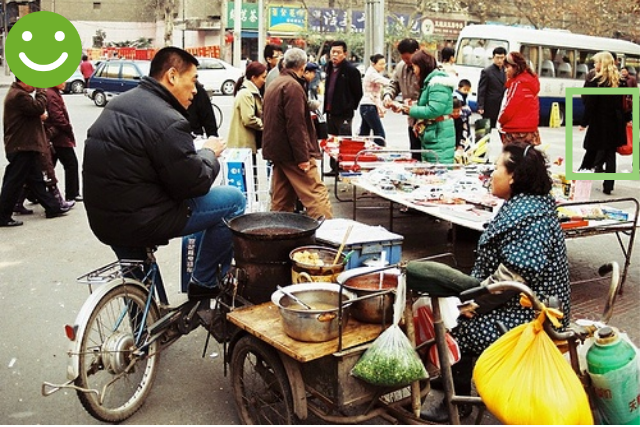} \\[6pt]
\rnk{3} & \includegraphics[width=\imgW,height=\imgH]{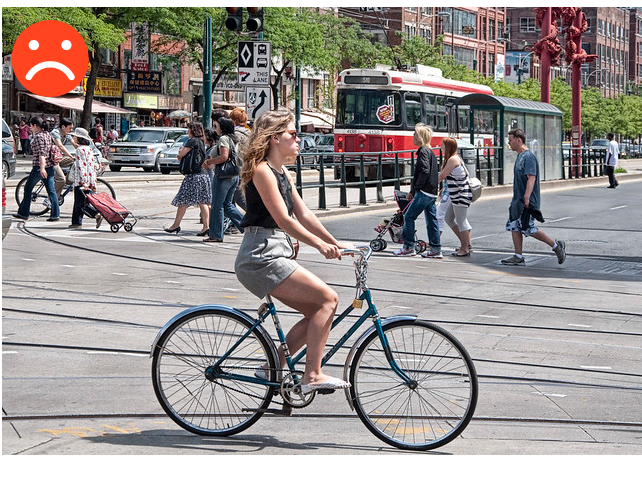} & \includegraphics[width=\imgW,height=\imgH]{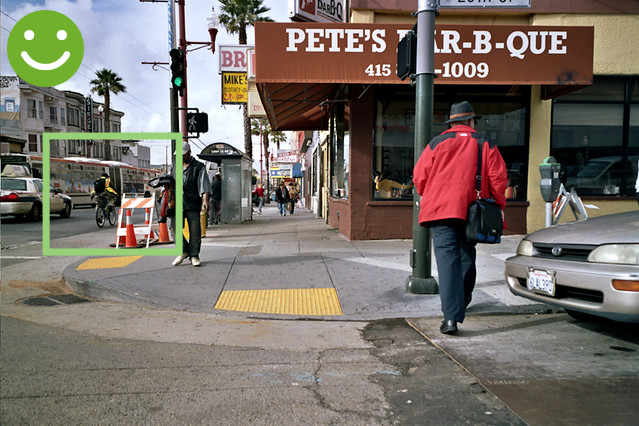} & \includegraphics[width=\imgW,height=\imgH]{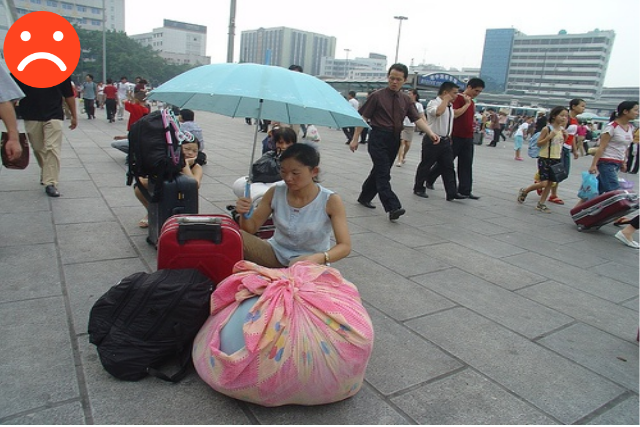} & \includegraphics[width=\imgW,height=\imgH]{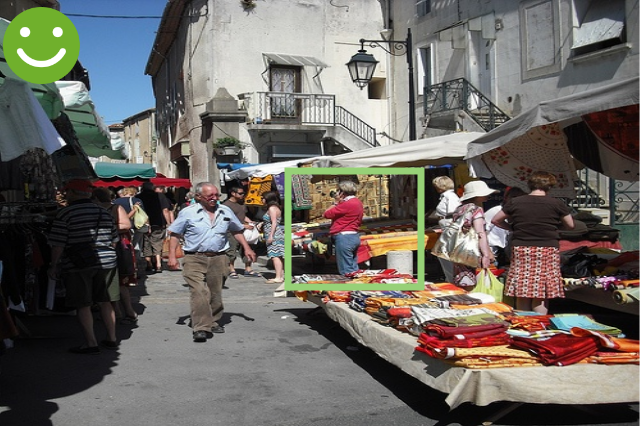} \\[6pt]
\rnk{4} & \includegraphics[width=\imgW,height=\imgH]{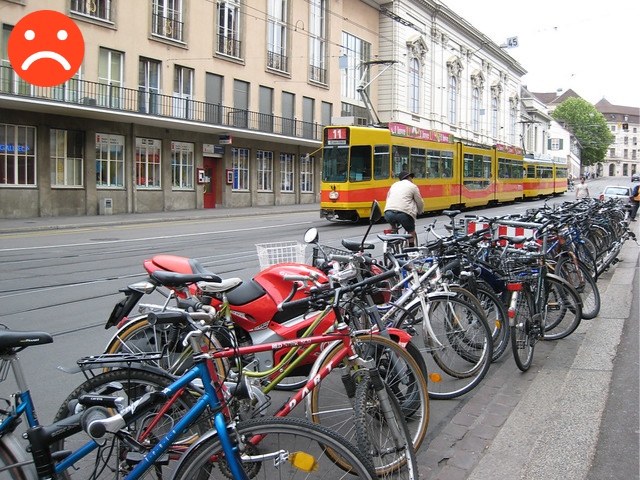} & \includegraphics[width=\imgW,height=\imgH]{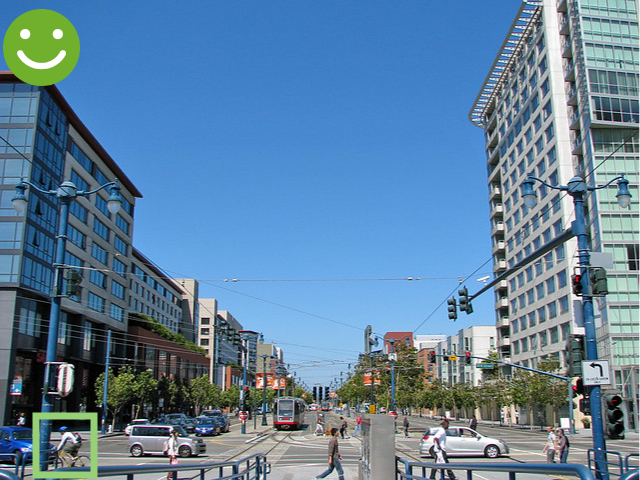} & \includegraphics[width=\imgW,height=\imgH]{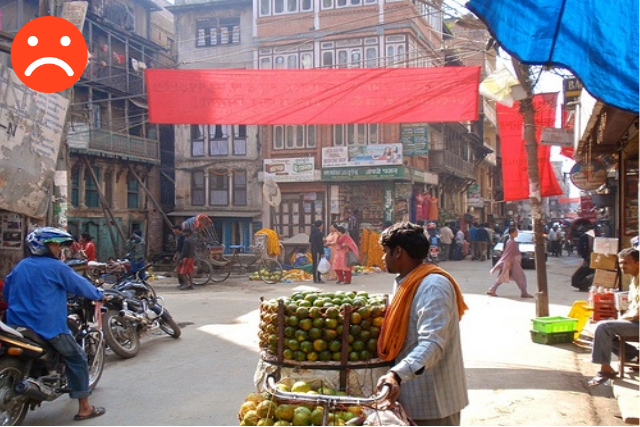} & \includegraphics[width=\imgW,height=\imgH]{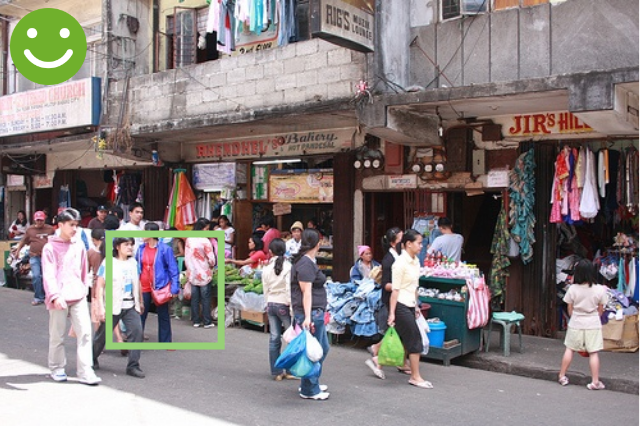} \\[6pt]
\rnk{5} & \includegraphics[width=\imgW,height=\imgH]{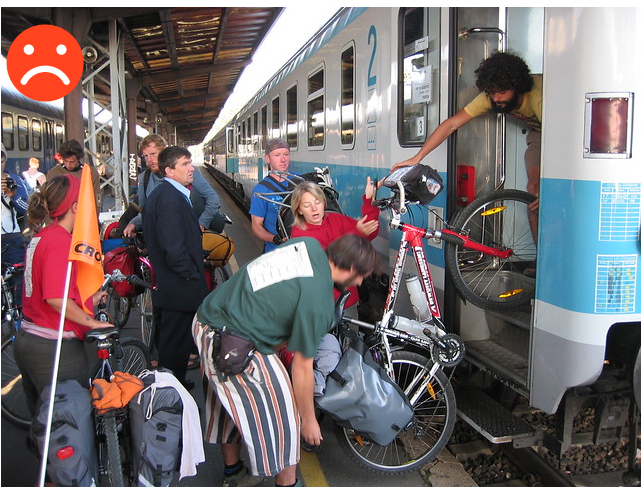} & \includegraphics[width=\imgW,height=\imgH]{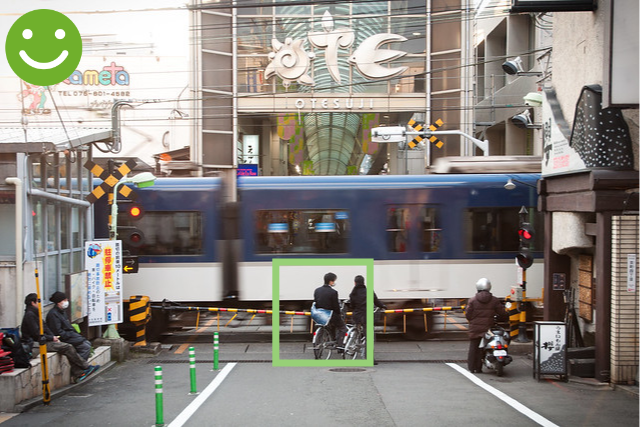} & \includegraphics[width=\imgW,height=\imgH]{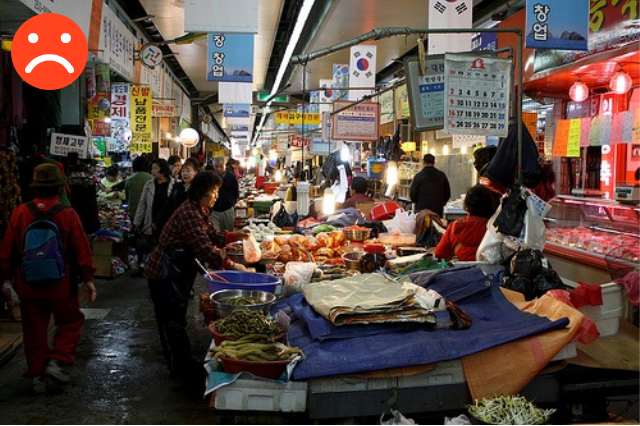} & \includegraphics[width=\imgW,height=\imgH]{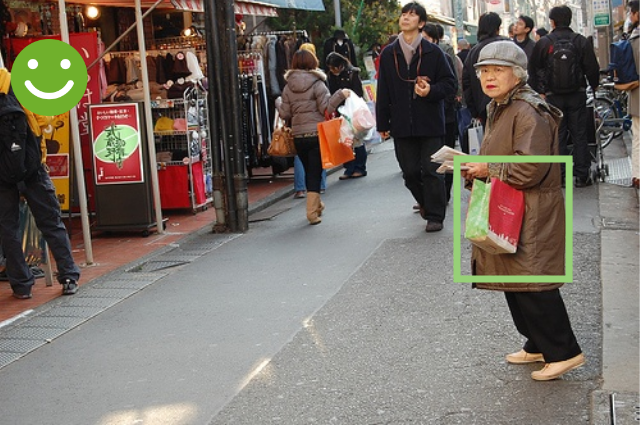} \\

\end{tabular}
\caption{\textbf{Qualitative comparison between Baseline and LARE} on COCO-Dense (Cols. 1--2) and Flickr30K-Dense (Cols. 3--4).
Top-5 retrieval results are shown; ground-truth is highlighted.
LARE improves ranking by leveraging fine-grained, localized cues missed by the baseline.}
\label{fig:retrieval_reranking}

\end{figure*}

\subsection{Qualitative Results}
\label{sec:qualitative}

Figure~\ref{fig:retrieval_reranking} presents qualitative comparisons between the baseline encoder (SigLIP) and LARE on dense retrieval queries from COCO-Dense (Columns 1--2) and Flickr30K-Dense (Columns 3--4). For each query, the top-5 retrieved images are shown, and the ground-truth image is highlighted with a dashed box.

In the first example (COCO-Dense), the query \textit{``A cyclist wearing a backpack next to a train station''} requires recognition of the backpack in addition to the cyclist and station context. The baseline ranks a generic cyclist at Rank~1, failing to capture the backpack attribute, while the correct image appears lower in the ranking. In contrast, LARE identifies the backpack as a localized discriminative cue and promotes the correct image to the top position for retrieval.

In the second example (Flickr30K-Dense), the query \textit{``A person carrying a red bag in a busy outdoor market''} hinges on detecting the red bag within a crowded scene. The baseline retrieves general market scenes that align with the global context but miss the specific attribute described in the query. LARE successfully retrieves the image containing the person with the red bag at Rank~1, indicating improved alignment with fine-grained details.

These examples illustrate that improvements arise when relevant evidence is spatially localized and visually subordinate within the scene. By incorporating region-level representations, LARE resolves ambiguities that global embeddings alone fail to distinguish. When global similarity is already reliable, rankings remain unchanged, consistent with the confidence-gated design.

\subsection{Comparison with Fine-Grained Methods}
\label{sec:finegrained}

Fine-grained alignment methods such as FILIP~\cite{yao2021filip}, RegionCLIP~\cite{zhong2021regionclip}, and ELIP~\cite{zhan2025elip} are not directly comparable to LARE, because each relies on training or query-time conditioning that a frozen, training-free pipeline does not provide. FILIP matches text tokens to image patches with a late-interaction score that it learns during pretraining; on a frozen encoder this score is not meaningful and retrieves at close to chance level, because the patch tokens are nearly orthogonal to the pooled embedding that the contrastive objective aligns with text. RegionCLIP retrains the encoder so that cropped regions align with text, whereas on a frozen encoder a tight crop drifts away from the contrastive space that retrieval depends on, while a larger crop that preserves context stays close to it. This is why LARE re-encodes spatially generous regions with the same frozen encoder rather than reusing patch tokens or tight crops. ELIP re-encodes each image conditioned on the query, which needs a trained prompting module and gives up the index-once property of large-scale retrieval, so it complements LARE rather than competing with it.

\subsection{Inference Overhead}
\label{sec:latency}

Because LARE encodes each image once globally and once per region, it raises the cost of building the retrieval index but not the cost of answering a query. Table~\ref{tab:latency} separates the two for SigLIP\,2 on a single GPU. Indexing is about six times more expensive than the baseline, as each image now needs six encoder passes instead of one; this cost is paid once, offline, and is amortized over all future queries, since the regional embeddings are computed when the index is built and stored alongside the global embedding.

\begin{table}[t]
\centering
\caption{Per-image and per-query cost of LARE on SigLIP\,2. The overhead falls on offline index building; query latency is unchanged.}
\label{tab:latency}
\footnotesize
\setlength{\tabcolsep}{6pt}
\begin{tabular}{lcc}
\toprule
 & \textbf{Baseline} & \textbf{LARE} \\
\midrule
Indexing (per image)  & 69\,ms    & 434\,ms \\
Retrieval (per query) & 4.7\,ms   & 5.0\,ms \\
Storage (per image)   & $1\times$ & $6\times$ \\
\bottomrule
\end{tabular}
\end{table}

At query time nothing changes. A query is still a single text encoding followed by one similarity computation, and the confidence gate only adjusts the score of uncertain pairs, so per-query latency matches the baseline. The practical price of LARE is therefore extra storage and a one-time, parallelizable indexing step, not slower retrieval. When indexing cost is itself a concern, the regional passes can be deferred to a re-ranking stage that crops only the top candidates of each query, leaving the index unchanged. Accuracy also saturates around five regions and degrades gracefully with fewer, so this budget can be lowered when needed.

\section{Conclusion}
\label{sec:conclusion}

We presented LARE, a training-free augmentation for text-to-image retrieval in crowded scenes. Our method mines low-attention regions from a frozen vision encoder, encodes these regions alongside the full image, and combines regional embeddings with the global image embedding at inference time. This simple test-time procedure improves retrieval on Dense-Set variants that emphasize subtle and occluded content.
We also introduced Dense-Set, a challenging crowded-scene benchmark derived from COCO and Flickr30K, where images are re-captioned to emphasize low attended areas. By shifting the focus toward fine-grained object, Dense-Set reveals the limitations of existing retrieval models and provides a more rigorous evaluation setting for densely crowded scenes.

For future work, we plan to make region selection more query-aware so that only the most informative crops are encoded, reducing compute while preserving accuracy gains. We also aim to strengthen fine-grained text--image alignment through patch-level interactions in the spirit of FILIP~\cite{yao2021filip}. In addition, extending LARE to temporal retrieval settings is a promising next step, building on dual-encoder video retrieval formulations such as CLIP4Clip and Frozen in Time~\cite{luo2022clip4clip,bain2021frozen}.
\bibliography{example_paper}
\bibliographystyle{icml2026}

\clearpage
\appendix
\graphicspath{{appendix/}{sec/fig/appendix/}}

\section{Additional Experimental Details}
\label{sec:appendix_details}

\subsection{Hyperparameter Sensitivity}
\label{sec:appendix_hyperparams}

We analyze the robustness of LARE with respect to its two primary inference-time hyperparameters: the number of selected regions $N$ and the confidence threshold $\tau$. These parameters control the balance between computational cost and retrieval refinement. Increasing $N$ allows the model to examine a broader set of candidate regions and improves the likelihood of recovering small or visually subtle objects that may be underrepresented in the global embedding. The threshold $\tau$ determines when regional refinement is activated, ensuring that additional computation is performed only when the global similarity signal is uncertain.

Overall, LARE remains stable across a wide range of settings and consistently improves retrieval performance over the baseline backbone. Performance increases as the number of regions grows, indicating that additional regional evidence helps resolve ambiguous queries. Beyond a moderate number of regions, gains saturate, suggesting that most relevant visual evidence has already been captured. Similarly, the method remains robust across different confidence thresholds. Based on this analysis, we use $N=5$ and $\tau=0.25$ throughout the paper, as this configuration provides a strong balance between retrieval accuracy and computational efficiency.

\FloatBarrier

\begin{figure*}[b]
\centering
\begin{minipage}[t]{0.485\textwidth}
    \centering
    \includegraphics[width=\linewidth]{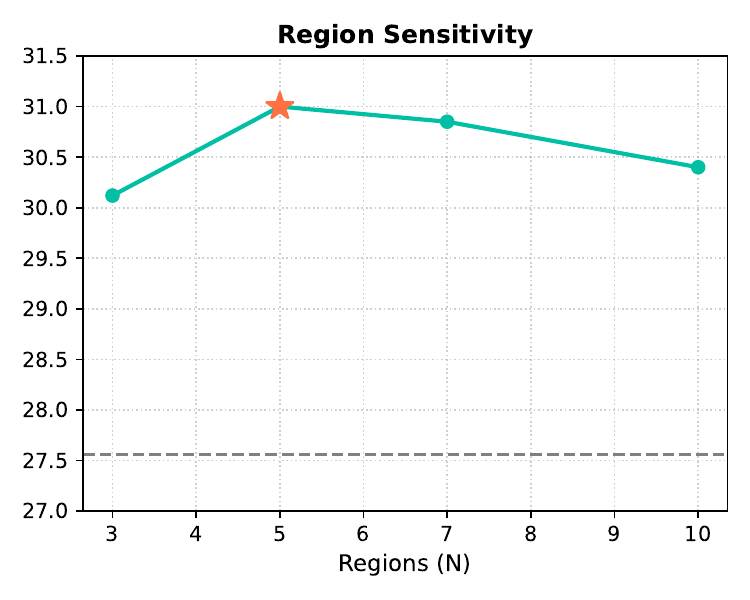}
    \par\small\textbf{(a)} Effect of region count $N$.
\end{minipage}
\hfill
\begin{minipage}[t]{0.485\textwidth}
    \centering
    \includegraphics[width=\linewidth]{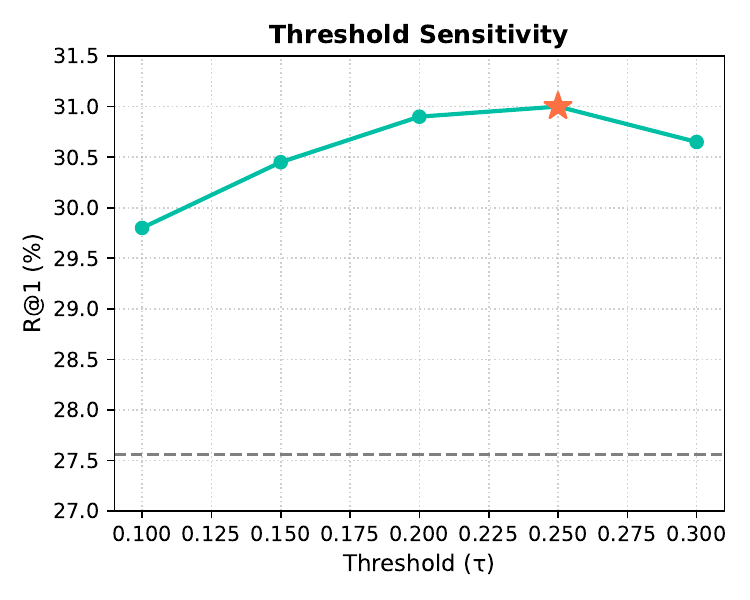}
    \par\small\textbf{(b)} Effect of confidence threshold $\tau$.
\end{minipage}

\vspace{0.8em}
\includegraphics[width=0.7\textwidth]{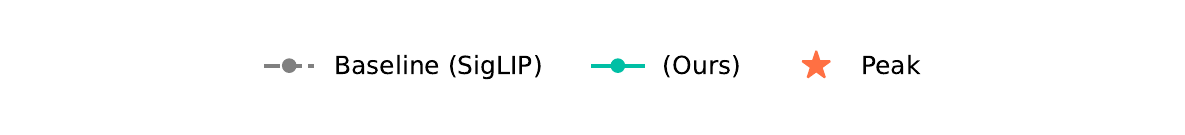}

\caption{
Sensitivity of LARE to inference hyperparameters.
Increasing the number of regions improves retrieval performance until saturation around $N=5$.
The method remains stable across thresholds and consistently outperforms the baseline.
}
\label{fig:hyperparam_sensitivity}
\vspace{-4mm}
\end{figure*}

\subsection{Implementation Notes}
\label{sec:appendix_impl_notes}

We follow the preprocessing and encoder configurations of the backbone models and use the OpenCLIP implementations~\cite{cherti2023reproducible} of CLIP and related ViT-based encoders. All encoders remain frozen, and LARE operates entirely at inference time without modifying model parameters or requiring additional training.

For each image, we extract the self-attention tensor from an intermediate transformer layer and compute the patch-to-patch attention maps (excluding the class token). For each head, we sum each column to measure how much attention a patch receives, reshape to a spatial grid, min--max normalize, and average the top-$k$ heads selected by spatial variance to obtain a mean attention map. We then form the inverse-attention map to identify regions that receive relatively low attention. Candidate regions are generated using a sliding window, merged using non-maximum suppression, and limited to at most $N$ regions. Each selected region is cropped from the original image, resized to the backbone's native input resolution, and encoded using the same frozen vision encoder to obtain regional embeddings.
During retrieval, LARE applies confidence-gated fusion: regional similarity is incorporated only when it provides stronger evidence than the global similarity score. This mechanism improves retrieval in dense scenes while preserving the original backbone behavior on standard benchmarks.
\clearpage
\section{Model Card}
\label{sec:model_card}

We provide a brief model card for LARE.

{\setlist[itemize]{leftmargin=*,itemsep=0.2em,topsep=0.2em,parsep=0pt,partopsep=0pt}
\begin{itemize}
    \item \textbf{Model Architecture}: LARE is a training-free augmentation pipeline that operates on frozen pretrained vision-language models. The pipeline contains three main components: (1) a vision transformer encoder for extracting global image embeddings and spatial attention maps, (2) a text transformer encoder for extracting text embeddings, and (3) an inverse-attention module that detects low-attention regions, re-encodes them independently, and adaptively fuses regional and global features. The vision and text encoders are frozen pretrained models, instantiated as CLIP ViT-L/14, SigLIP SoViT-400M/14, or SigLIP 2 SoViT-400M/16, accessed via OpenCLIP~\cite{cherti2023reproducible}.

    \item \textbf{Inputs}: The vision encoder takes an image as input, preprocessed to match the backbone's native resolution: $224 \times 224 \times 3$ for CLIP ViT-L/14, and $384 \times 384 \times 3$ for SigLIP and SigLIP 2 models. The text encoder takes a tokenized text string, cropped to the first 64 tokens as input.

    \item \textbf{Outputs}: The vision and text encoders output a $d$-dimensional feature vector, where $d$ is 768 for CLIP ViT-L/14 and 1152 for SigLIP and SigLIP 2 SoViT-400M models. The pipeline outputs a fused similarity score between the text query and image.

    \item \textbf{Intended Use}: The method is designed for zero-shot image--text retrieval research purposes. The pipeline can be used for text-to-image and image-to-text retrieval by comparing feature vectors. The method is particularly effective for challenging retrieval scenarios where queries target fine-grained details, small objects, or background elements that may be under-emphasized by global embeddings.

    \item \textbf{Training Data}: LARE requires no training or fine-tuning. All vision and text encoders are frozen pretrained models (e.g., CLIP and SigLIP). The inverse-attention module operates entirely at inference time and requires no additional training data.

    \item \textbf{Evaluation Data}: Zero-shot retrieval is performed on MS-COCO, Flickr30k, and a curated dense-scene dataset (Dense-Set) to demonstrate performance across different retrieval difficulty levels.

    \item \textbf{Hardware \& Software}: The method is implemented in Python using PyTorch and OpenCLIP and evaluated on NVIDIA Quadro RTX 8000 GPUs (48GB).
\end{itemize}
}
\section{Pseudocode}
\label{sec:appendix_pseudocode}

\begin{algorithm}[H]
\small
\caption{LARE: Low-Attention Region Encoding for Retrieval}
\label{alg:pipeline}
\begin{algorithmic}[1]
\REQUIRE Image $I$, text query $q$, frozen vision encoder $f_v$, text encoder $f_t$, layer $\ell$, top heads $k$, max regions $N$, confidence threshold $\tau$
\ENSURE Retrieval score $S$
\STATE \textit{\textbf{Stage 1: Low-Attention Region Detection}}
\STATE $\{\mathbf{A}^{(h)}\}_{h=1}^{H} \gets f_v(I, \ell)$ \COMMENT{Extract attention maps at layer $\ell$}
\FOR{each head $h = 1, \ldots, H$}
    \STATE $\mathbf{a}^{(h)}_i \gets \sum_j \mathbf{A}^{(h)}_{j,i}$ for all patches $i$ \COMMENT{Received attention}
    \STATE $\mathbf{a}^{(h)} \gets \textsc{MinMaxNorm}(\mathbf{a}^{(h)})$
\ENDFOR
\STATE $\mathcal{H}_k \gets$ top-$k$ heads by $\text{Var}(\mathbf{a}^{(h)})$
\STATE $\bar{\mathbf{A}} \gets \frac{1}{k} \sum_{h \in \mathcal{H}_k} \mathbf{a}^{(h)}$
\STATE $\mathbf{M} \gets \mathbf{1} - \bar{\mathbf{A}}$ \COMMENT{Inverse attention map}
\STATE $\mathcal{W} \gets \textsc{SlidingWindow}(\mathbf{M})$ \COMMENT{Candidate windows}
\STATE $\mathcal{R} \gets \textsc{NMS}(\mathcal{W})$
\STATE $\mathcal{R} \gets \textsc{TopN}(\mathcal{R}, N)$ \COMMENT{Keep top-$N$ regions}
\STATE \textit{\textbf{Stage 2: Regional Encoding}}
\FOR{each region $r_j \in \mathcal{R}$}
    \STATE $\mathbf{z}_j \gets f_v(\textsc{CropAndResize}(I, r_j))$
\ENDFOR
\STATE \textit{\textbf{Stage 3: Confidence-Gated Scoring}}
\STATE $\mathbf{z}_g \gets f_v(I)$; $\mathbf{z}_t \gets f_t(q)$
\STATE $s_g \gets \text{sim}(\mathbf{z}_t, \mathbf{z}_g)$ \COMMENT{Global similarity}
\STATE $s_r \gets \max_{j} \text{sim}(\mathbf{z}_t, \mathbf{z}_j)$ \COMMENT{Best regional match}
\IF{$s_g < \tau$ \textbf{and} $s_r > s_g$}
    \STATE $\alpha \gets \min\bigl(2(s_r - s_g),\, 0.5\bigr)$
    \STATE $S \gets (1 - \alpha)\, s_g + \alpha\, s_r$
\ELSE
    \STATE $S \gets s_g$
\ENDIF
\STATE \textbf{return} $S$
\end{algorithmic}
\end{algorithm}

\end{document}